\documentclass[10pt,twocolumn,letterpaper]{article}

 \usepackage[pagenumbers]{cvpr} %

\definecolor{cvprblue}{rgb}{0.21,0.49,0.74}
\usepackage[pagebackref,breaklinks,colorlinks,allcolors=cvprblue]{hyperref}

\usepackage{booktabs}
\usepackage{multirow}
\usepackage{multicol}
\usepackage{color, colortbl}

\def\paperID{6684} %
\def\confName{CVPR}
\def\confYear{2025}

\title{Doe-1: Closed-Loop Autonomous Driving with Large World Model}

\author{
 Wenzhao Zheng{\footnotemark[1] $^,$\footnotemark[2]}\quad  Zetian Xia\footnotemark[1]\quad Yuanhui Huang\quad Sicheng Zuo\quad  Jie Zhou\quad Jiwen Lu \\
Department of Automation, Tsinghua University, China \\
\texttt{ wenzhao.zheng@outlook.com; xiazt21@mails.tsinghua.edu.cn} \\
Project Page: \url{https://wzzheng.net/Doe} \\
Large Driving Models: \url{https:/github.com/wzzheng/LDM}
}

\begin{document}

\twocolumn[{%
\renewcommand\twocolumn[1][]{#1}%
\vspace{-5mm}
\maketitle
\vspace{-10mm}
\begin{center}
    \centering
    \includegraphics[width=1\linewidth]{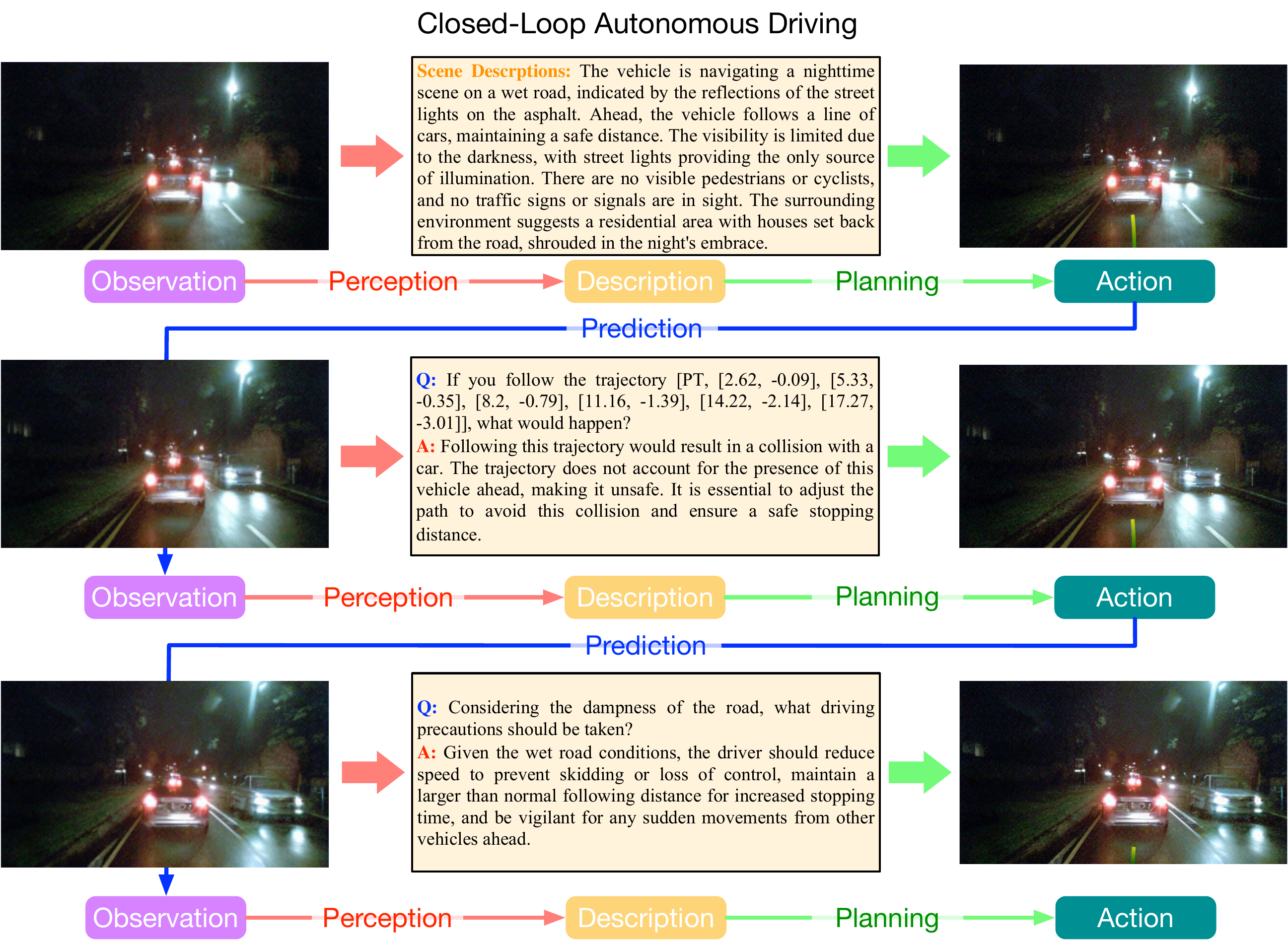}
    \vspace{-7mm}
    \captionof{figure}{    
    \textbf{Visualizations of our \textbf{Doe-1} for closed-loop autonomous driving on nuScenes~\cite{nuscenes}.}
We propose a large driving world model (\textbf{Doe-1}) to achieve unified generative closed-loop autonomous driving.
We model perception, prediction, and planning as the transitions of observation$\rightarrow$description, description$\rightarrow$action, and action$\rightarrow$observation, respectively. 
\textbf{Doe-1} accomplishes perception, planning, and prediction in a unified autoregressive generative framework and achieves closed-loop end-to-end autonomous driving for the first time.
}
\label{fig:supp_teaser}
\end{center}%
}]

\renewcommand{\thefootnote}{\fnsymbol{footnote}}
\footnotetext[1]{Equal contribution. $\dagger$Project leader.}
\renewcommand{\thefootnote}{\arabic{footnote}}

\begin{abstract}
End-to-end autonomous driving has received increasing attention due to its potential to learn from large amounts of data.
However, most existing methods are still open-loop and suffer from weak scalability, lack of high-order interactions, and inefficient decision-making.
In this paper, we explore a closed-loop framework for autonomous driving and propose a large \textbf{D}riving w\textbf{O}rld mod\textbf{E}l (\textbf{Doe-1}) for unified perception, prediction, and planning.
We formulate autonomous driving as a next-token generation problem and use multi-modal tokens to accomplish different tasks. 
Specifically, we use free-form texts (i.e., scene descriptions) for perception and generate future predictions directly in the RGB space with image tokens. 
For planning, we employ a position-aware tokenizer to effectively encode action into discrete tokens.
We train a multi-modal transformer to autoregressively generate perception, prediction, and planning tokens in an end-to-end and unified manner.
Experiments on the widely used nuScenes dataset demonstrate the effectiveness of \textbf{Doe-1} in various tasks including visual question-answering, action-conditioned video generation, and motion planning. 
Code: \url{https://github.com/wzzheng/Doe}.
\end{abstract}
    
\section{Introduction}
\label{sec:intro}

The emergence of GPT series~\cite{gpt,gpt2,gpt3} stimulates the rapid development of large models with various functions, including language modeling~\cite{llama, llama2, qwen}, visual undertanding~\cite{llava,llava2,qwenvl,gpt4}, and decision-making~\cite{openvla,rt2,gpt_driver}.  
The key to the success of large models is scaling up model sizes and training data~\cite{kaplan2020scaling}.
When designing a model, the scalability advocates large representation compacity (e.g., transformers~\cite{vaswani2017attention, vit, swin}) over well-designed inductive biases (e.g., convolution neural networks~\cite{resnet,densenet}) to improve the upper bound of performance.

To build large models for autonomous driving, some methods directly apply large language models (LLMs)~\cite{llmassist,drivingllm,languagempc,gpt_driver} or vision-language models (VLMs)~\cite{lmdrive, drivegpt4,drivecot,drivevlm,drivemlm,carllava,adh} for motion planning~\cite{drivevlm} or scene question-answering~\cite{reason2drive, omnidrive}. 
They usually align the inputs with texts (e.g., Q-Former~\cite{blip2}) and output language descriptions of the planning results~\cite{gpt_driver}.
However, LLMs are known to share the hallucination issue~\cite{liu2024survey,gunjal2024detecting,li2023evaluating}, hindering the interpretability and safety of autonomous driving.
To avoid this, others follow the well-tested pipeline of perception, prediction, and planning for autonomous driving and explore a scalable end-to-end model~\cite{hu2022stp3,uniad,vad,ye2023fusionad,genad,tong2023scene} to jointly accomplish them.
Though promising, most existing methods are still open-loop and suffer from several issues.
\textbf{1) Weak scalability.} They use manually designed scene representation which cannot provide comprehensive information for downstream tasks.
\textbf{2) Lack of high-order interactions.} They predict future scenes without considering the planned ego trajectory.
\textbf{3) Inefficient decision-making.} They usually plan several steps ahead yet practically only use the first step to execute.

\begin{figure*}[t]
\centering
\includegraphics[width=\textwidth]{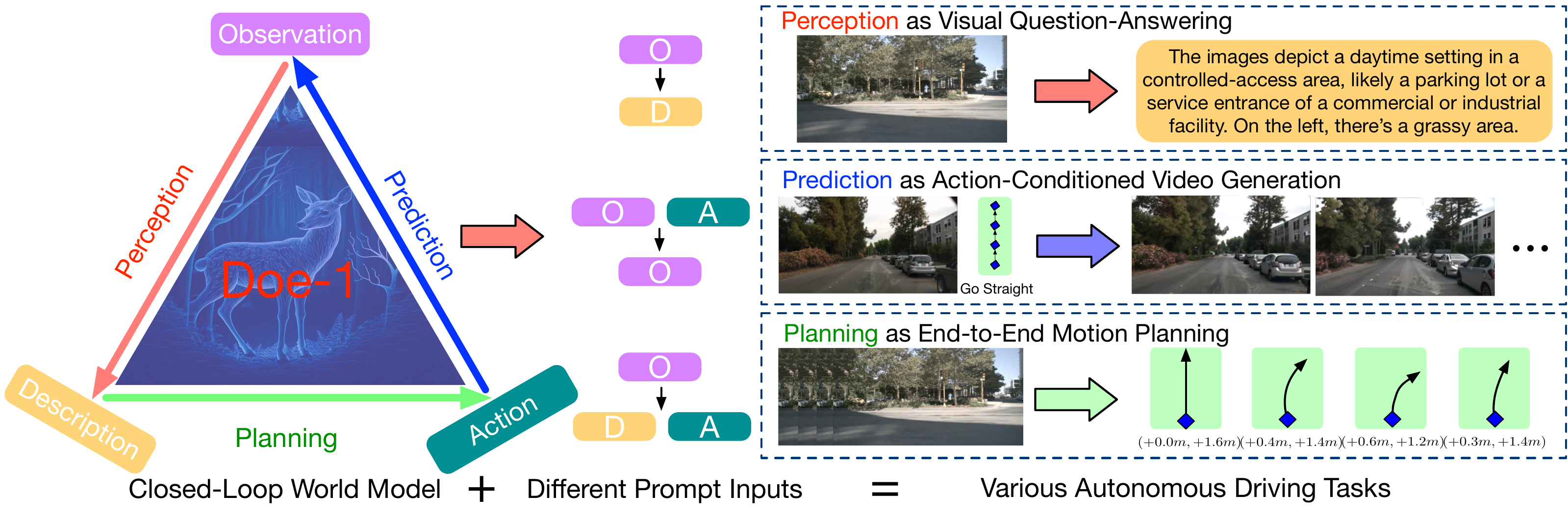}
\vspace{-9mm}
    \captionof{figure}{
    \textbf{Overview of the proposed \textbf{Doe-1}.}
    We formulate autonomous driving as a unified next-token generation problem and use observation, description, and action tokens to represent each scene.
    Without additional fine-tuning, \textbf{Doe-1} accomplishes various tasks by using different input prompts, including visual question-answering, controlled image generation, and end-to-end motion planning.
    }
\label{teaser}
\vspace{-6mm}
\end{figure*}

To address thesex, we propose a closed-loop large \textbf{D}riving w\textbf{O}rld mod\textbf{E}l (\textbf{Doe-1}) for unified perception, prediction, and planning without intermediate latent scene representations, as shown in Figure~\ref{teaser}.
We cast autonomous driving as a scene evolution problem and represent each scene with observation, description, and action tokens.
We then formulate the conventional perception, planning, anaction-conditioned video generationd prediction as transitions between multi-modal tokens, i.e., observation$\rightarrow$description, description$\rightarrow$action, and action$\rightarrow$observation, respectively. 
We then employ a generative autoregressive world model to model this evolution with next-token prediction.
Concretely, we focus on vision-centric autonomous driving and adopt RGB images as observations.
We tokenize images using the image vector-quantized variational autoencoder~\cite{oord2017neural}.
We employ free-form texts as the scene descriptions and also feed the model with question-answering pairs for on-demand perception.
We represent action with displacement in the bird's eye view (BEV) space and employ a position-aware tokenizer to encode it into discrete tokens. 
Our \textbf{Doe-1} then generates the observation, description, and action tokens of the next temporal frame sequentially and iteratively, which can be efficiently trained with a simple autoregressive objective.
We conduct extensive experiments on the nuScenes dataset~\cite{nuscenes} to evaluate the capability of our \textbf{Doe-1}.
With different prompt settings, we demonstrate that our \textbf{Doe-1} successfully achieves various tasks without finetuning, including visual question-answering, controlled image generation, and motion planning.

\section{Related Work}
\label{sec:formatting}
\textbf{Large Models for Autonomous Driving.}
The success of GPT series~\cite{gpt,gpt2,gpt3} confirms the power of model and data scaling and promotes the rapid development of large language models (LLMs)~\cite{llama, llama2, qwen}, which demonstrate impressive performance on a wide range of language tasks. 
Early methods combining LLMs with autonomous driving use ground-truth labels~\cite{languagempc,drivingllm,llmassist} or pre-trained perception models~\cite{gpt_driver} to obtain detection and map results.
They then translate them into text descriptions as inputs and leverage the reasoning ability of LLMs to output the planned trajectory in text.
To enable LLMs to process images, large vision-language models (VLMs)~\cite{llava,llava2,qwenvl,gpt4} usually pre-align the image representations with text space and then jointly finetune the overall model.
This facilitates the emergence of end-to-end planning methods which take as inputs images and directly output the planned trajectory.
Further methods explore the use of chain-of-thoughts~\cite{drivecot,reason2drive}, instruction tuning~\cite{instructdriver}, or additional visual question answering~\cite{omnidrive} to improve the interpretability. 
Still, existing methods struggle with the hallucination issue~\cite{liu2024survey,gunjal2024detecting,li2023evaluating} commonly possessed by LLMs and VLMs and thus lack robustness and safety, which are critical for autonomous driving.
To alleviate this issue, we follow the well-tested perception, prediction, and planning pipeline and propose a large driving world model (\textbf{Doe-1}) to unify all these tasks for more robust autonomous driving.

\textbf{World Models for Autonomous Driving.} 
The conventional world model aims to predict the next observation of the scene given the action of the ego agent~\cite{sutton1991dyna, ha2018world}.
Most existing methods follow this definition and focus on controlled generation in the image~\cite{panacea,yang2024generalized,vista,delphi,stag-1}, LiDAR~\cite{khurana2022eo,khurana2023point,mersch2022self,weng2021inverting} or 3D occupancy~\cite{occworld,occsora} space.
Vision-based methods usually utilize the capability of pre-trained image diffusion models (e.g., StableDiffusion~\cite{rombach2022high}) and finetune them on the driving scenarios to predict action-conditioned futures~\cite{gao2023magicdrive, drivedreamer,yang2023bevcontrol,li2023drivingdiffusion}.
The other methods explore more diverse architectures (e.g., autoregressive transformer~\cite{hu2023gaia,occworld,gpd-1}, gated recurrent unit~\cite{muvo}, or autoencoder~\cite{agro2024uno, yang2024driving}) to generate future scenes. 
However, most methods cannot be directly applied to trajectory planning and are usually used to compute rewards to train another planner with reinforcement learning~\cite{vista}.
OccWorld~\cite{occworld} generalizes the concept of world model and predicts the joint evolution of 3D occupancy and the ego vehicle.
Still, it requires pre-trained perception models to obtain 3D occupancy and is limited by its representation ability.
Differently, the proposed \textbf{Doe-1} employs one closed-loop unified framework for perception, prediction, and planning as a generalized autonomous driving model.

\textbf{End-to-End Autonomous Driving.}
Perception, prediction, and planning are the well-tested pipeline for autonomous driving and are conventionally realized with separate modules~\cite{tpvformer,phan2020covernet,scheel2022urban, cheng2022mpnp}.
Recent methods explore the end-to-end design that jointly trains all the modules with a shared objective of trajectory planning~\cite{hu2022stp3,uniad,vad,ye2023fusionad,genad,tong2023scene}.
The unified framework equips them with more feasibility to scale up and more potential to achieve native large models for autonomous driving. 
Early methods~\cite{hu2022stp3,uniad} adopt bird's eye view as the scene representation to pass information throughout the model.
Subsequent methods explore the use of 3D occupancy~\cite{occworld,tong2023scene} for more details or sparse queries~\cite{sparsead,sparsedrive} for better efficiency.
However, the manually designed scene representations might not be optimal and the induced inductive biases limit the capability of the model, which is critical for scale-up.
They also suffer from the lack of high-order interactions and inefficient decision-making.
In this paper, we propose the first closed-loop autonomous driving model, \textbf{Doe-1}, for more scalable autonomous driving, which achieves perception with free-form texts and predicts futures directly in the image space to reduce information loss and enlarge the model capacity.

\section{Proposed Approach}

\subsection{Closed-Loop Autonomous Driving}

Autonomous driving is a long-standing task to apply artificial intelligence to the real world, which aims to plan the future actions $\{ \mathbf{a} \}$ for the ego vehicle from scene observations $\{ \mathbf{o} \}$.
Several attempts~\cite{bojarski2016end,codevilla2018end,prakash2020exploring,chen2020learning,chitta2022transfuser} explore the use of deep neural networks to directly model the mapping from observations $\{ \mathbf{o} \}$ to actions $\{ \mathbf{a} \}$:
\begin{equation}\label{eqn: ad}
\begin{aligned}
\{ \mathbf{o}^{t-H}, \cdots, \mathbf{o}^{t} \} \xrightarrow{} \{ \mathbf{a}^{t}, \cdots, \mathbf{a}^{t+F-1} \},
\end{aligned}
\end{equation}
where $t$ is the current time stamp, and $H$ and $F$ is the number of concerned history and future frames, respectively.
The black-box nature of deep neural networks makes the decision-making process less transparent and trustworthy.

Recent methods utilize the reasoning ability of large language models (LLMs) or vision-language models (VLMs) to improve the interpretability of the planning results~\cite{instructdriver,reason2drive,languagempc,drivingllm,llmassist}.
They align the visual features with the text space by direct transcription~\cite{languagempc} or learnable projection~\cite{lmdrive}.
Some methods further use additional text descriptions $\mathbf{d}$ as an auxiliary task~\cite{omnidrive} to further alleviate the hallucination issue~\cite{liu2024survey} of large models:
\begin{equation}\label{eqn: vlm}
\begin{aligned}
\{ \mathbf{o}^{t-H}, \cdots, \mathbf{o}^{t} \} \xrightarrow{\text{LLM/VLM}} \mathbf{d}^t, \{ \mathbf{a}^{t}, \cdots, \mathbf{a}^{t+F-1} \}.
\end{aligned}
\end{equation}
Still, it is doubtful whether LLMs/VLMs truly understand the 4D dynamics and interactions among traffic agents without fine-grained understanding and prediction of the scene.

\begin{figure}[t]
\centering
\includegraphics[width=0.475\textwidth]{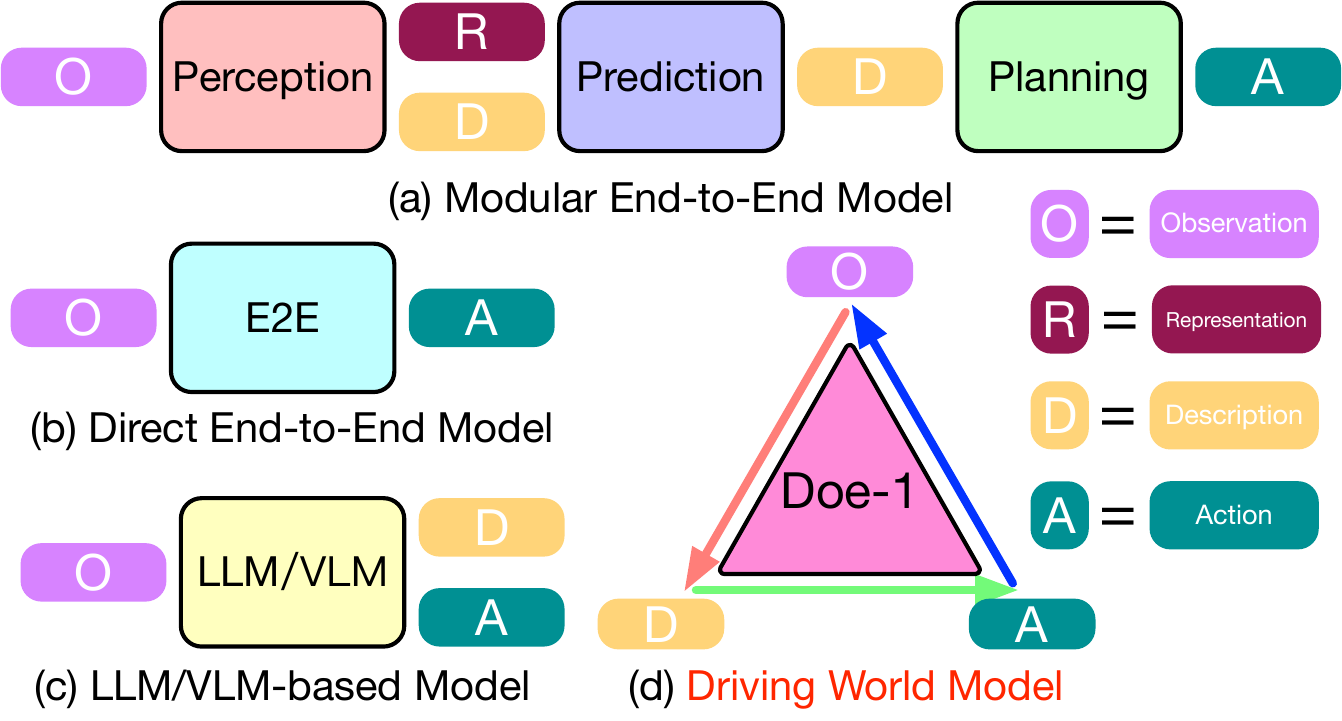}
\vspace{-7mm}
\caption{\textbf{Comparisons of different paradigms.}
(a) The modular end-to-end model performs perception, prediction, and planning sequentially and is the most popular pipeline for autonomous driving.
(b) The direct end-to-end model directly outputs the planned action given sensor inputs.
(c) The LLM/VLM-based model exploits the reasoning ability of LLMs/VLMs to output actions.
(d) The proposed driving world model (\textbf{Doe-1}) predicts the evolutions between observations, descriptions, and actions to achieve close-loop end-to-end autonomous driving.
}
\label{fig:comparison}
\vspace{-6mm}
\end{figure}

To build more trustworthy autonomous driving systems, the mainstream methods (including modern end-to-end models~\cite{hu2022stp3,uniad,vad,ye2023fusionad,genad,tong2023scene}) employ a ``divide and conquer'' strategy and decompose autonomous driving into sequential perception $f$, prediction $g$, and planning $h$:
\begin{equation}\label{eqn: e2e}
\begin{aligned}
& \{ \mathbf{o}^{t-H}, \cdots, \mathbf{o}^{t} \} \xrightarrow{f_1} \mathbf{r}^t \xrightarrow{f_2} \mathbf{d}^t \xrightarrow{g} \\
& \{ \mathbf{d}^{t+1}, \cdots, \mathbf{d}^{t+F} \} \xrightarrow{h} \{ \mathbf{a}^{t}, \cdots, \mathbf{a}^{t+F-1} \},
\end{aligned}
\end{equation}
where $\mathbf{r}$ and $\mathbf{d}$ denotes the scene representation and scene description, respectively.
The scene representation $\mathbf{r}$ is a set of continuous features in the 3D space (e.g., bird's eye view~\cite{bevdepth,bevformer}, voxels~\cite{surroundocc}, tri-perspective view~\cite{tpvformer,pointocc}, 3D Gaussians~\cite{gaussianformer,gaussianformer-2,embodiedocc}, dense points~\cite{driv3r}),
while $\mathbf{d}$ provides high-level refined descriptions (e.g., detected bounding boxes, constructed map) of the scene.
The design of $\mathbf{d}$ introduces prior knowledge of the most important factors for decision-making and thus can improve the robustness of the model.
We compare different paradigms in Figure~\ref{fig:comparison}.

Despite its satisfactory performance, we argue that this well-known pipeline still suffers from several issues including weak scalability, lack of high-order interactions, and inefficient decision-making.

\textbf{Weak Scalability.}
One advantage of end-to-end driving is its potential to scale up model sizes and fit more training data~\cite{occworld}.
However, most existing methods use manually designed scene representations $\mathbf{r}$ and descriptions $\mathbf{d}$, which induce inductive bias and prior knowledge.
Though this could be beneficial with little available training data, it limits the performance upper bound when scaling up.
For example, some factors (e.g., the pose of a human, the rear light of a vehicle) can be crucial in certain corner cases, and discarding them will lead to inaccurate decisions.

\begin{figure}[t]
\centering
\includegraphics[width=0.475\textwidth]{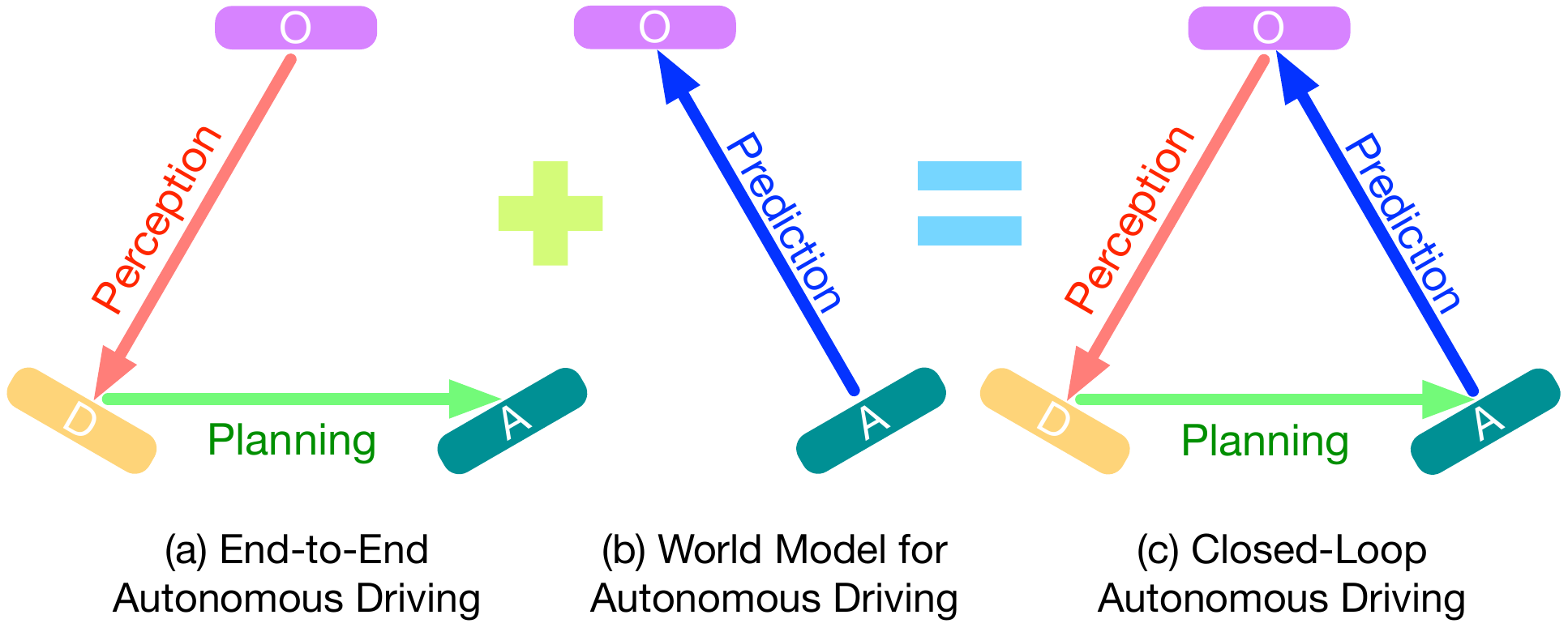}
\vspace{-7mm}
\caption{\textbf{Illustration of the proposed closed-loop autonomous driving paradigm.}
(a) Existing end-to-end autonomous driving methods (e.g., UniAD~\cite{uniad}, GenAD~\cite{genad}) usually perform perception first and then make decisions according to the perceived descriptions.
(b) Existing world models for autonomous driving (e.g., DriveDreamer~\cite{drivedreamer}, OccWorld~\cite{occworld}) predict future observations based on the current actions.
(c) Close-loop autonomous driving combines the two paradigms to construct a closed loop.
}
\label{fig:closed-loop}
\vspace{-5.5mm}
\end{figure}

\textbf{Lack of High-Order Interactions.}
The current pipeline \eqref{eqn: e2e} usually predicts the future $F$ scenes at once before planning, assuming that the ego action \textbf{Doe-1}s not affect the environment development.
However, in the interactive scenarios of autonomous driving, the future movements of other agents are highly dependent on the ego action.
For example, vehicles behind usually will respond to the deceleration of the ego car.
Therefore, it is not feasible to accurately predict the future without conditions of the ego action.

\textbf{Inefficient Decision-Making.}
Most existing methods predict and plan several steps ahead, as stated in \eqref{eqn: e2e}.
However, in practice, driving models only take the planned next-frame action and re-plan future actions given new observations, resulting in inefficiency and redundancy of multi-step planning.
Therefore, we think the model needs to be capable of predicting possible futures for possible ego actions but plan the instant (i.e., next-frame) action without explicit deduction.
For example, we humans have the ability to predict how the environment respond to our actions, yet we usually make decisions instinctly and do not explicitly ponder the chain of reactions when driving.

To address these issues, this paper explores a new closed-loop autonomous driving paradigm, as shown in Figure~\ref{fig:closed-loop} and proposes a large driving world model, \textbf{Doe-1}, to formulate autonomous driving as an autoregressive world evolution of multimodal states:
\begin{equation}\label{eqn: Doe}
\begin{aligned}
\mathbf{o}^t \xrightarrow{\text{\textbf{Doe-1}}} \mathbf{d}^t \xrightarrow{\text{\textbf{Doe-1}}} \mathbf{a}^{t} \xrightarrow{\text{\textbf{Doe-1}}} \mathbf{o}^{t+1} \xrightarrow{\text{\textbf{Doe-1}}} \mathbf{d}^{t+1} \xrightarrow{\text{\textbf{Doe-1}}} \cdots.
\end{aligned}
\end{equation}
Using self-attention, \textbf{Doe-1} directly accesses the observations and predicts future scene evolutions from the observation space without intermediate scene representations.
\textbf{Doe-1} makes predictions with ego action as the condition.
Through autoregressive generation, \textbf{Doe-1} predicts multi-step futures yet only makes instant planning each time.

\begin{figure*}[t]
\centering
\includegraphics[width=1\textwidth]{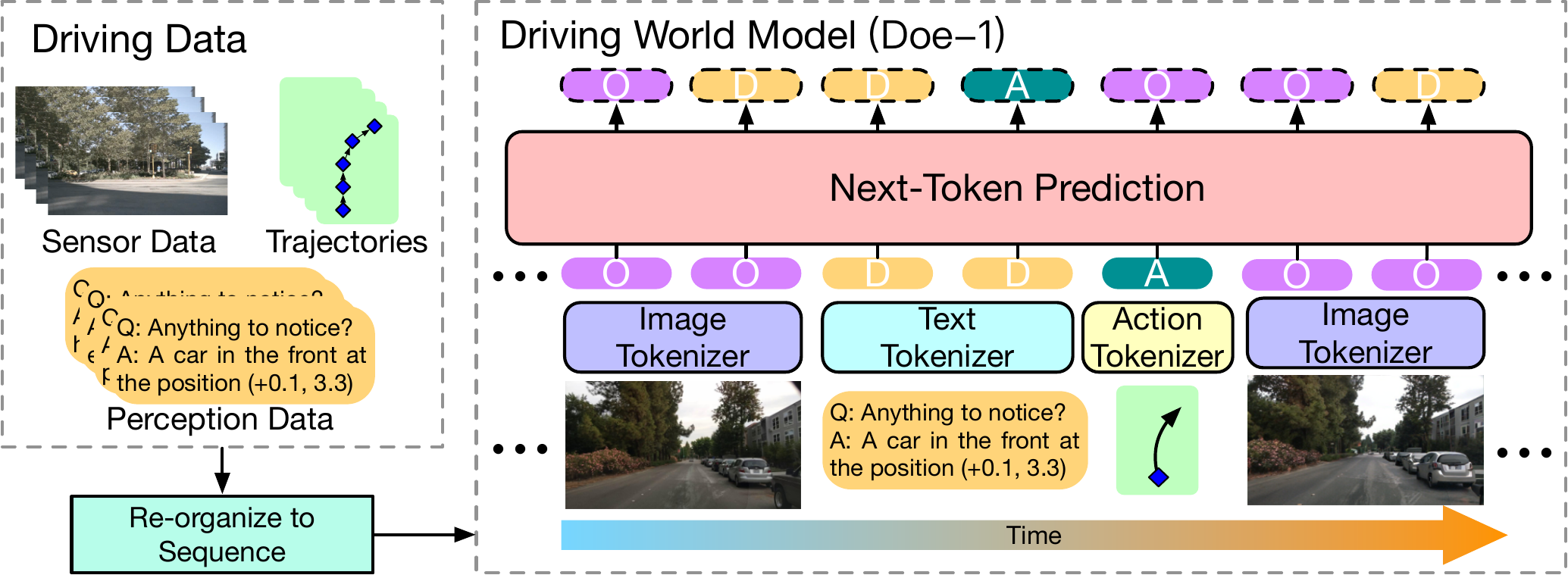}
\vspace{-7mm}
\caption{\textbf{Framework of the proposed \textbf{Doe-1}.}
We first re-organize the training dataset into a temporal sequence of sensor data (image), perception data (texts), and action data (position of the next frame). 
We then use image, text, and action tokenizers to encode them into discrete tokens to construct a 1D token sequence.
We then use a transformer-based architecture to autoregressively model this sequence and use the next-token prediction as the training objective.
}
\label{fig:framework}
\vspace{-6mm}
\end{figure*}

\subsection{\textbf{Doe-1}: A Large Driving World Model}

\textbf{Overview.}
The overall framework of \textbf{Doe-1} is depicted in Figure~\ref{fig:framework}. We use Chameleon architecture~\cite{team2024chameleon} as a unified multimodal autoregressive model to generate predictions. Following \ref{eqn: Doe}, the observations $\mathbf{o} \in \mathbb{R}^{3\times h \times w}$, descriptions $\mathbf{d} \in \mathbb{Z}^l$ and actions $\mathbf{a} \in \mathbb{R}^{p\times3}$ are tokenized into discrete space and simply concatenated into sequences, where $(h,w)$ is the shape of observation images, $l$ is the length of string $\mathbf{d}$ and $p$ is the number of motions making $\mathbf{a}$ contains $p$ motion parameters in 2D space. Using the past sequences as prompt, the unified model will predicted the next token iteratively. Specifically, given the tokenized observations $\mathbf{T}_o^{i} \in \mathbb{Z}^{d_{o}}$, the high-level descriptions $\mathbf{T}_d^{i} \in \mathbb{Z}^{d_d}$ and the actions $\mathbf{T}_a^{i} \in \mathbb{Z}^{d_a}$ from time $i$ where $0 \le i \le t$, the model gives predictions tokens at time $t+1$ as follows:
\begin{equation}\label{eqn: autoregress}
\begin{aligned}
& \mathbf{T}_o^{t+1} = \underset{\mathbf{T}_o}{\arg\max} \ p(\mathbf{T}_o|\mathbf{T}_o^{0},\mathbf{T}_d^{0},\mathbf{T}_a^{0} ,\cdots, \mathbf{T}_a^t), \\
& \mathbf{T}_d^{t+1} = \underset{\mathbf{T}_d}{\arg\max} \ p(\mathbf{T}_d|\mathbf{T}_o^{0},\mathbf{T}_d^{0},\mathbf{T}_a^{0} ,\cdots, \mathbf{T}_a^t,\mathbf{T}_o^{t+1}), \\
& \mathbf{T}_a^{t+1} = \underset{\mathbf{T}_a}{\arg\max} \ p(\mathbf{T}_a|\mathbf{T}_o^{0},\mathbf{T}_d^{0},\mathbf{T}_a^{0} ,\cdots, \mathbf{T}_o^{t+1},\mathbf{T}_d^{t+1}).
\end{aligned}
\end{equation}

The $\mathbf{o}_{t+1}$, $\mathbf{d}_{t+1}$, $\mathbf{a}_{t+1}$ are then decoded from the predicted tokens. To encode the sequence $\{\mathbf{o}_i,\mathbf{d}_i,\mathbf{a}_i\}_{i=0}^t$ into discrete tokens, an observation tokenizer, a description tokenizer and an action tokenizer are used.

\textbf{Observation Tokenzier.}
Observations are images in the RGB space $\mathbb{R}^{3\times h\times w}$. To map the image $\mathbf{o}$ to the tokens $\mathbf{T}_o$ in a discrete space, we use the pretrained image tokenizer from Lumina-mGPT~\cite{liu2024lumina}, which encodes an image into 1024 discrete tokens from a codebook of size 8192 and adds special tokens to indicate the shape of the image.

\textbf{Description Tokenzier.}
In interactive scenarios, a description $\mathbf{d}$ are in a char space $\mathbb{Z}^l$, where l denotes the length of string $\mathbf{d}$. We use the BPE tokenizer from Chameleon~\cite{team2024chameleon} to encode texts of descriptions into a discrete space.

Treating floating-point numbers as text results in the loss of their original metric properties in the Euclidean space. For the floating-point number sequences in the texts of descriptions (such as the distance to an object from the ego), we round them to a resolution of 0.02 meters, then the discretized floating point numbers are mapped to 4000 discrete tokens in turn. In order to maintain the original metric properties of numbers after embedding, we use the Sinusoidal Positional Encoding as the embedded features of these tokens to ensure that floating-point numbers with small differences still maintain a small distance after being embedded in the high-dimensional space. For a floating-point $x\in \mathbb{R}$, the token $T$ and the embedding of $T$ is computed as:
\begin{equation}\label{eqn: pe}
\begin{aligned}
&T(x) = \text{round}(\frac{1}{r}\times x)+b, \\ 
&\text{Embedding}(x, 2i) = \sin(\frac{T(x)}{\text{scale}^{2i/d}}), \\
&\text{Embedding}(x, 2i+1) = \cos(\frac{T(x)}{\text{scale}^{(2i+1)/d}}),
\end{aligned}
\end{equation}
where $r$ denotes the resolution, $b$ denotes a bias constant determined by the codebook size and $scale$ is a scaling constant determined by the range of $x$. The positional embedding is set to be non-learnable to ensure that the floating-point numbers which are not seen in the training phase are still aligned in the embedding space.

\textbf{Action Tokenzier.}
The motion $\mathbf{z}$ denotes the movement of a vehicle between two frames and is defined as a three-element array $\mathbf{z}=(d_x,d_y,\theta)$ containing displacements along two horizontal axes and a yaw angle, and an action $\mathbf{a} \in \mathbb{R}^{p\times 3}$ given by the model concatenates $p$ consecutive moves to denote the future trajectory in the next $p$ frames. To encode actions into discrete tokens, we use positional encoding similar to above and simply adjust the scaling constant. Special tokens are inserted as anchors at the beginning and end of an action.

\textbf{Generative Architecture.}
By aligning the multimodal generation, we can readily scale up the unified model. 
For any modality, the model follows a unified autoregressive paradigm. When predicting actions, the model generates several motion predictions for multiple future frames at once by giving $p(\mathbf{a}_t|\mathbf{T})$ based on the previous token sequence $\mathbf{T}$, and $\mathbf{a_t}$ will affect the prediction of subsequent tokens. In order to avoid the accumulation of errors between predictions at different frames, last $p-1$ motions of the generated predictions are masked when predictions $\mathbf{a}\in \mathbb{R}^{p\times 3}$ are generated, with the first prediction retained to provide a reference for generating subsequent observations:
\begin{equation}\label{eqn: mask}
\begin{aligned}
& \mathbf{a}_t=(\mathbf{z}_t,\mathbf{z}_{t+1},\cdots,\mathbf{z}_{t+p}), \\
& \text{masked}(\mathbf{a}_t)=(\mathbf{z}_t,\mathbf{m}_1,\cdots,\mathbf{m}_p),
\end{aligned}
\end{equation}
where $\mathbf{m}_i$ denotes the $i$-th mask array which will be encoded into mask tokens.

\subsection{Applications of \textbf{Doe-1}}

\begin{figure}[t]
\centering
\includegraphics[width=0.475\textwidth]{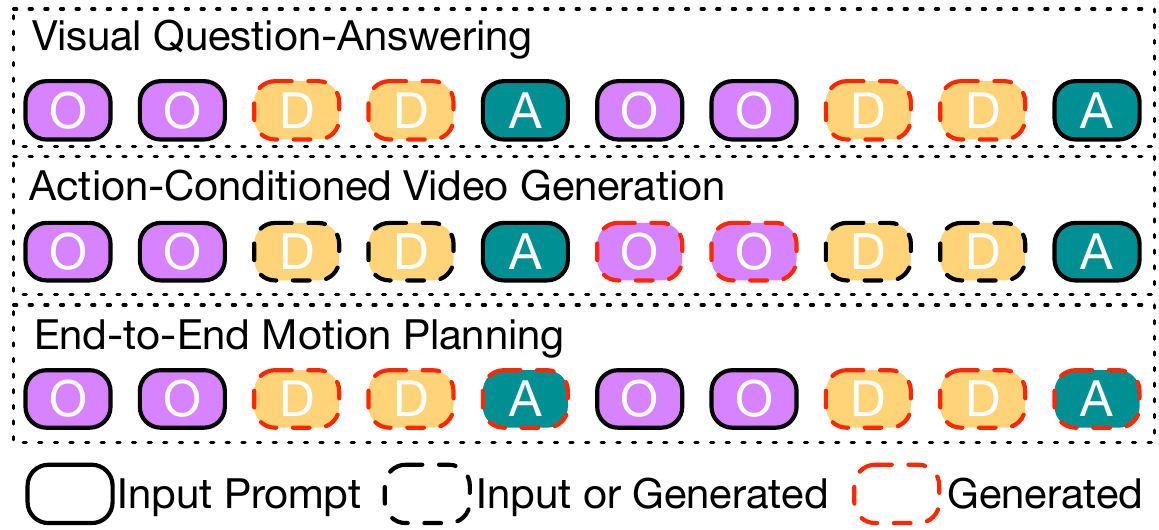}
\vspace{-7mm}
\caption{\textbf{Illustration of different prompt settings to apply our \textbf{Doe-1} to different tasks.}
For visual question-answering, we use observation tokens as prompt input and generate description tokens.
For action-conditioned video generation, we use the observation and action tokens of the current frame as inputs and generate observation tokens of the next frame.
For end-to-end motion planning, we use observation tokens as prompts to generate the description and action tokens.
}
\label{fig:application}
\vspace{-6mm}
\end{figure}

As a unified multimodal world model, \textbf{Doe-1} supports inputs from multiple modalities and automatically predicts the next modality, facilitating the applications to different tasks simply by altering the prompt. Figure~\ref{fig:application} introduces the application of \textbf{Doe-1} to different tasks of visual question-answering, motion planning, and action-conditioned video generations as examples. By designing prompt sequences, \textbf{Doe-1} can be transferred to other multimodal tasks.

\textbf{Visual Question-Answering.}
Given the observation $\mathbf{o}$, the model is required to generate a precise description $\mathbf{d_o}$ of the observation. Furthermore, to assess the model's understanding of the scene and the perception ability , we require the model to complete interactive question-answering tasks based on the given observations and the generated descriptions. These tasks include identifying objects in the scene that may impact driving, describing the environment, and checking the plausibility of a given driving trajectory, among others, which requires the model to have as comprehensive a description capability of the scene as possible.

\begin{table*}[t]
\caption{\textbf{Visual question-answering results on OmniDrive-nuScenes~\cite{omnidrive}.}
$^*$ denotes only using the front camera as input.
}
\vspace{-3mm}
\centering
\setlength{\tabcolsep}{12pt}
\begin{tabular}{l|cc|ccc}
\toprule
\multirow{2}{*}{Method} &
\multicolumn{2}{c|}{Counterfactual} &
\multicolumn{3}{c}{Caption} \\
& AP (\%) $\uparrow$ & AR (\%) $\uparrow$ & METEOR$\uparrow$ & CIDEr $\uparrow$ & ROUGE$\uparrow$\\
\midrule
OmniDrive-3D~\cite{omnidrive} & {52.3} & \textbf{59.6}  & 38.0 & 68.6 & 32.6 \\
\midrule
OmniDrive-2D~\cite{omnidrive} & 51.5 & 55.0 & \textbf{38.3} & 67.1 & 32.5\\
OmniDrive-BEV~\cite{omnidrive} & 45.6 & 49.5 & 35.6 & 59.5 & 27.8 \\
\midrule 
\textbf{\textbf{Doe-1}}$^*$ & \textbf{54.5} & 54.0 & 37.6 & \textbf{72.8} & \textbf{35.6} \\
\bottomrule
\end{tabular}
\vspace{-2mm}
\label{vqa}
\end{table*}

\begin{figure*}[t]
\centering
\includegraphics[width=\textwidth]{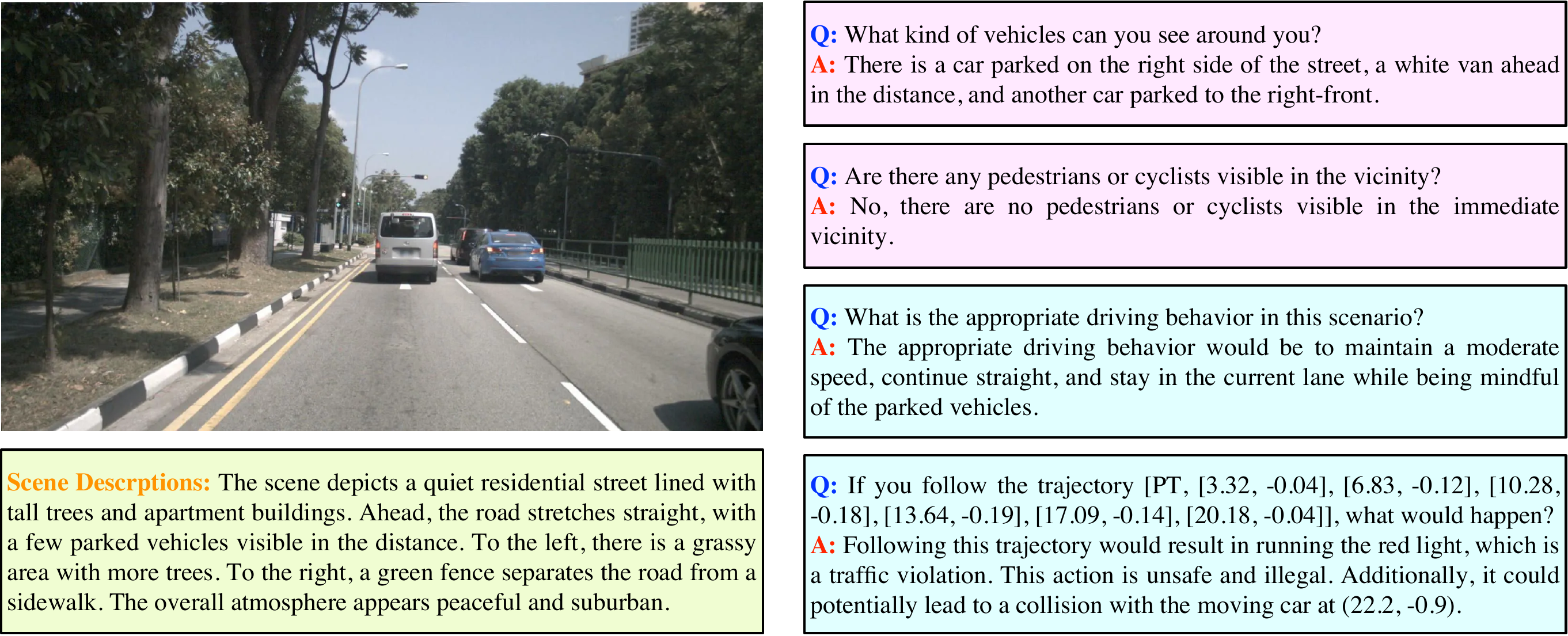}
\vspace{-7mm}
\caption{\textbf{Visualizations of the visual-question answering results of our \textbf{Doe-1}.}
\textbf{Doe-1} produces language descriptions and answers questions about the scene.
}
\label{fig:vis_vqa}
\vspace{-6mm}
\end{figure*}

\begin{table*}[t]
\setlength{\tabcolsep}{1.5pt}
\caption{\textbf{Action-conditioned video generation results on the nuScenes~\cite{caesar2020nuscenes} dataset.}
}
\vspace{-3mm}
\centering
\begin{tabular}{l|ccccccc}
\toprule
\multirow{1}{*}{{Method}}
& DriveGAN~\cite{drivegan} & DriveDreamer~\cite{drivedreamer} & WoVoGen~\cite{wovogen} & Drive-WM~\cite{wang2023driving} & GenAD~\cite{yang2024generalized} & Vista~\cite{vista} & \textbf{\textbf{Doe-1}} \\
\midrule
{Type} & GAN & Diffusion & Diffusion & Diffusion & Diffusion & Diffusion & Auto-Regressive \\
{Resolution} & 256 $\times$ 256 & 128 $\times$ 192 & 256 $\times$ 448 & 192 $\times$ 384 & 256 $\times$ 448 & 576 $\times$ 1024 & 384 $\times$ 672 \\
\hline
{FID $\downarrow$} & 73.4 & 52.6 & 27.6 & 15.8 & 15.4 & \textbf{6.9} & 15.9 \\
\bottomrule
\end{tabular}
\vspace{-3mm}
\label{gen}
\end{table*}

\begin{figure*}[t]
\centering
\includegraphics[width=\textwidth]{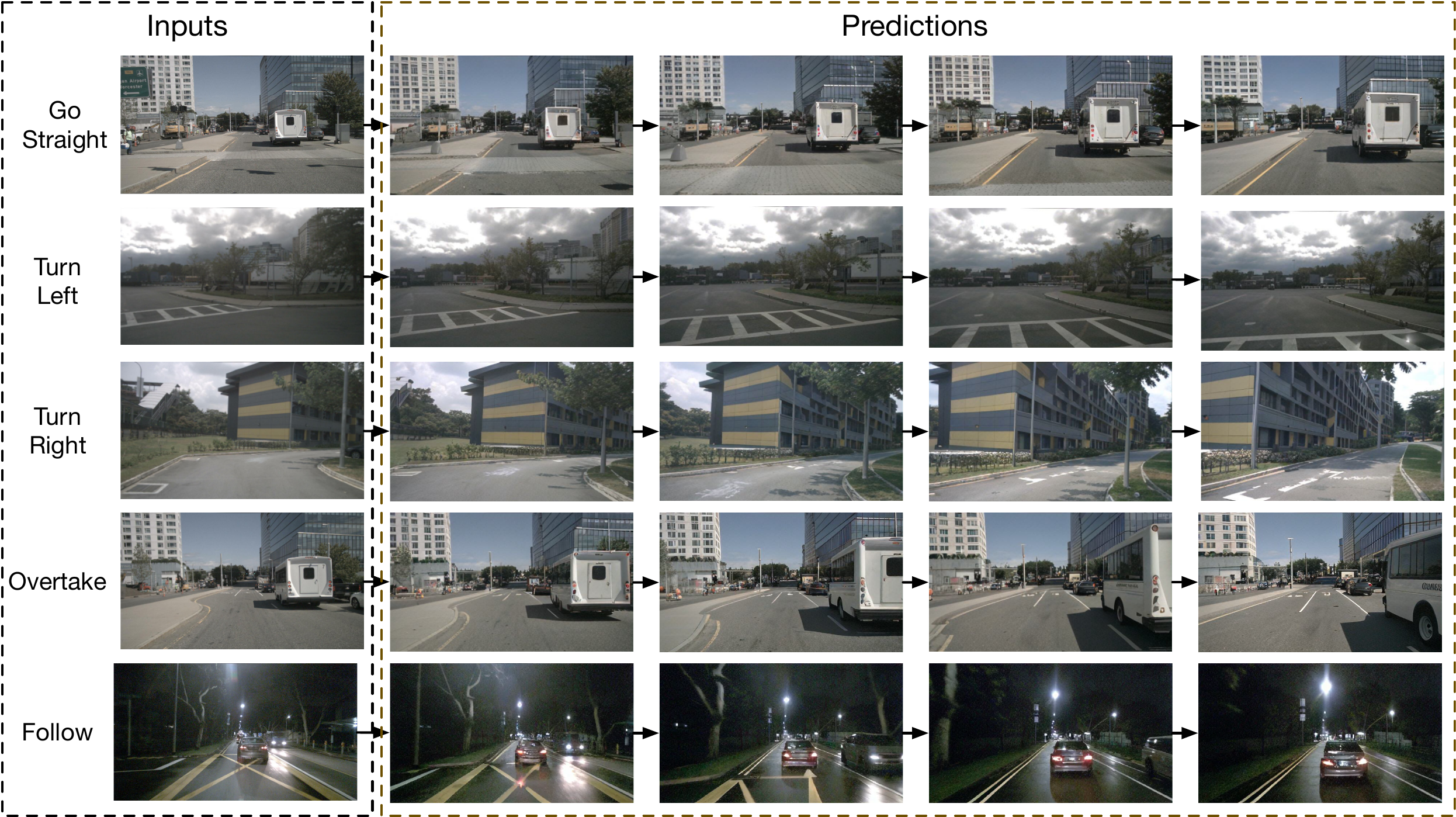}
\vspace{-7mm}
\caption{\textbf{Visualizations of action-conditioned video generation results of our \textbf{Doe-1}.}
\textbf{Doe-1} generates high-quality videos in consistent with the 3D structures and action conditions.
}
\label{fig:vis_gen}
\vspace{-5mm}
\end{figure*}

\textbf{Action-Conditioned Video Generation.}
We hope that the world model, \textbf{Doe-1}, can enable closed-loop interactive simulation of driving scenarios without the need for feedback from real-world interactions. This requires \textbf{Doe-1} to be able to predict the changes in observations after taking an action. Given a sequence of observations with descriptions $\{\mathbf{o}_j,\mathbf{d}_j\}_{j=0}^t$ and the actions history $\{\mathbf{a_j}\}_{j=0}^t$, the model is required to generate the observations $\mathbf{o}_{t+1}$.

In the action-conditioned video generation task, for the observations, we only provide the model with the image at time step 0, and the model is required to iteratively generate the next p frames of images based on the given actions.

\textbf{Motion Planning.}
Given a sequence of observations $\{\mathbf{o}_j\}_{j=0}^t$ and history motions $\{\mathbf{z}_j\}_{j=0}^{t-1}$, the model generates descriptions $\mathbf{d_t}$ and plan future motions $\{\mathbf{z}_{t+i}\}_{i=0}^p$ efficiently, where $p$ denotes the number of predicted frames.

By combining the aforementioned three tasks, the model can independently perform closed-loop driving simulation.

\section{Experiments}

In this section, we evaluate the performance of our model on a variety of driving-related tasks. Despite only using a single-view camera as input and high-level QAs as supervision, the model still exhibits promising performance.
\subsection{Dataset}
We conducted experiments on the widely use nuScenes dataset~\cite{caesar2020nuscenes} for autonomous driving.
It includes 1,000 driving sequences captured in diverse scenarios, including daytime/nighttime and sunny/cloudy/rainy weather.
Each video clip consists of 400 frames with a frequency of 20Hz, spanning across 20 seconds as a scene.
They are downsampled to 2Hz to obtain keyframes and annotated with labels including scene descriptions, 3D object bounding boxes, and semantic maps.
Though it provides point clouds scanned by a 32-beam LiDAR, we mainly focus on using the RGB images captured by six surrounding cameras as inputs.
We follow existing methods~\cite{genad} to use the training 700 scenes for training and the validation 150 scenes for testing.

\subsection{Task Descriptions}
We evaluate our \textbf{Doe-1} on various tasks including visual question-answering, motion planning, and action-conditioned video generation.

\textbf{Visual Question-Answering.}
We evaluate the perception ability of our method on the OmniDrive-nuScenes~\cite{omnidrive} benchmark, which supplements the original nuScenes data with high-quality visual question-answering (QA) text pairs generated by GPT4.
The QA pairs include perception, reasoning, and planning in the 3D space.
The goal of visual QA is to generate correct answers to questions about the visual observations.
We use the widely used language-based metrics (i.e., {METEOR}~\cite{banerjee2005meteor}, {ROUGE}~\cite{lin2004rouge}, and {CIDEr}~\cite{vedantam2015cider}) to compute the sentence similarities between the generated and ground-truth answers.
For counterfactual reasoning, we ask the model to predict what will happen if executing a given trajectory.
Following OmniDrive~\cite{omnidrive}, we extract keywords from the predictions, including ``safety'', ``collision'', ``running a red light'', and ``out of the drivable area''.
We compare them with the ground truth and compute the average precision (AP) and recall (AR) for all categories.

\textbf{Action-Conditioned Video Generation.}
The objective of action-conditioned video generation is to generate high-quality images consistent with the controlling conditions. 
World models in autonomous driving are usually formulated as generating future observations given the actions (e.g., high-level command, trajectory).
Following existing methods~\cite{vista,drivedreamer}, we report the Fr\'echet Inception Distance (FID)~\cite{ho2020denoising} score to measure the image generation quality.

\textbf{End-to-End Motion Planning.}
End-to-end motion planning aims to produce safe and consistent future ego trajectories from observations.
 To assess performance, we use L2 error and collision rate as the key evaluation metrics.
We evaluate the performance of our GaussianAD with the L2 error and collision rate following existing end-to-end methods~\cite{hu2022stp3, uniad, genad, occworld}.
The L2 error computes the L2 distance between the planned and ground-truth paths.
The collision rate reflects how often the autonomous vehicle collides with other objects while executing the planned trajectory. 
For fair comparisons, we use a 5-frame (i.e., 2 seconds) history and compute the metrics for the 1s, 2s, and 3s future. 

\begin{table*}[t]
\setlength{\tabcolsep}{0.01\linewidth}
\caption{\textbf{End-to-end motion planning results on the nuScenes~\cite{caesar2020nuscenes} dataset.} 
$^*$ represents only using the front camera as input, and $\dagger$ denotes using the metric adopted in VAD~\cite{vad}. }
\vspace{-3mm}
\begin{tabular}{l|lc|cccc|cccc}
\toprule
\multirow{2}{*}{Method} & \multirow{2}{*}{Input} & \multirow{2}{*}{Auxiliary Supervision} &
\multicolumn{4}{c|}{L2 (m) $\downarrow$} & 
\multicolumn{4}{c}{Collision Rate (\%) $\downarrow$} \\
&& & 1s & 2s & 3s & \cellcolor{gray!30}Avg. & 1s & 2s & 3s & \cellcolor{gray!30}Avg.  \\
\midrule
IL~\cite{ratliff2006maximum} & LiDAR & None  & 0.44 & 1.15 & 2.47 & \cellcolor{gray!30}1.35 & 0.08 & 0.27 & 1.95 & \cellcolor{gray!30}0.77  \\
NMP~\cite{zeng2019nmp} & LiDAR & Box \& Motion & 0.53 & 1.25 & 2.67 & \cellcolor{gray!30}1.48 & 0.04 & 0.12 & 0.87 & \cellcolor{gray!30}0.34  \\
FF~\cite{hu2021ff} & LiDAR & Freespace  & 0.55 & 1.20 & 2.54 & \cellcolor{gray!30}1.43 & 0.06 & 0.17 & 1.07 & \cellcolor{gray!30}0.43  \\
EO~\cite{khurana2022eo} & LiDAR & Freespace  & 0.67 & 1.36 & 2.78 & \cellcolor{gray!30}1.60 & 0.04 & 0.09 & 0.88 & \cellcolor{gray!30}0.33  \\
\midrule
\midrule
ST-P3~\cite{hu2022stp3} & Camera & Map \& Box \& Depth & 1.33 & 2.11 & 2.90 & \cellcolor{gray!30}2.11 & 0.23 & 0.62 & 1.27 & \cellcolor{gray!30}0.71  \\
UniAD~\cite{uniad} & Camera & { \footnotesize Map \& Box \& Motion \& Tracklets \& Occ}  & {0.48} & {0.96} & {1.65} & \cellcolor{gray!30}{1.03} & {0.05} & {\textbf{0.17}} & \textbf{0.71} & \cellcolor{gray!30}{\textbf{0.31}}  \\
OccNet~\cite{tong2023scene} & Camera & 3D-Occ \& Map \& Box & 1.29 & 2.13 & 2.99 & \cellcolor{gray!30}2.14 & 0.21 & 0.59 & 1.37 & \cellcolor{gray!30}0.72  \\
OccWorld~\cite{occworld} & Camera & 3D-Occ & 0.52 & 1.27 & 2.41 & \cellcolor{gray!30}1.40 & 0.12 & 0.40 & 2.08 & \cellcolor{gray!30}0.87  \\
VAD-Tiny~\cite{vad}  & Camera & Map \& Box \& Motion  & 0.60 & 1.23 & 2.06 & \cellcolor{gray!30}1.30 & 0.31 & 0.53 & 1.33 & \cellcolor{gray!30}0.72  \\
VAD-Base~\cite{vad} & Camera & Map \& Box \& Motion & 0.54 & 1.15 & 1.98 & \cellcolor{gray!30}1.22 & {\textbf{0.04}} & 0.39 & 1.17 & \cellcolor{gray!30}0.53 \\
GenAD~\cite{genad} & Camera & Map \& Box \& Motion & {\textbf{0.36}} & {\textbf{0.83}} & {\textbf{1.55}} & \cellcolor{gray!30}{\textbf{0.91}} & 0.06 & {0.23} & {1.00} & \cellcolor{gray!30}{0.43} \\
\midrule
\textbf{\textbf{Doe-1}} & Camera$^*$ & QA & 0.50 & 1.18 & 2.11 & \cellcolor{gray!30}1.26 & \textbf{0.04} & 0.37 & 1.19 & \cellcolor{gray!30}0.53  \\
\midrule
\midrule
\color{gray}VAD-Tiny$^\dagger$~\cite{vad}  & \color{gray}Camera & \color{gray}Map \& Box \& Motion  & \color{gray}0.46 & \color{gray}0.76 & \color{gray}1.12 & \color{gray}\cellcolor{gray!30}0.78 & \color{gray}0.21 & \color{gray}0.35 & \color{gray}0.58 & \color{gray}\cellcolor{gray!30}0.38  \\
\color{gray}VAD-Base$^\dagger$~\cite{vad} & \color{gray}Camera & \color{gray}Map \& Box \& Motion & \color{gray}0.41 & \color{gray}0.70 & \color{gray}1.05 & \color{gray}\cellcolor{gray!30}0.72 & \color{gray}{0.07} & \color{gray}0.17 & \color{gray}0.41 & \color{gray}\cellcolor{gray!30}0.22 \\
\color{gray}{OccWorld-D}$^\dagger$~\cite{occworld} & \color{gray}Camera & \color{gray}3D-Occ & \color{gray}{0.39} & \color{gray}0.73 & \color{gray}1.18 & \color{gray}\cellcolor{gray!30} 0.77 & \color{gray}0.11 & \color{gray}{0.19} & \color{gray}0.67 & \color{gray}\cellcolor{gray!30} 0.32  \\
\color{gray}GenAD$^\dagger$~\cite{genad} & \color{gray}Camera & \color{gray}Map \& Box \& Motion & \color{gray}0.28 & \color{gray}0.49 & \color{gray}0.78 & \color{gray}\cellcolor{gray!30}0.52 & \color{gray}0.08 & \color{gray}{0.14} & \color{gray}\textbf{0.34} & \color{gray}\cellcolor{gray!30}{\textbf{0.19}}  \\
\color{gray}{OmniDrive}$^\ddagger$~\cite{omnidrive} & \color{gray}Camera & \color{gray}Map \& Box \& Motion \&  QA & \color{gray}\textbf{0.14} & \color{gray}\textbf{0.29} & \color{gray}\textbf{0.55} & \cellcolor{gray!30}\color{gray}\textbf{0.33} & \color{gray}\textbf{0.00} & \color{gray}\textbf{0.13} & \color{gray}0.78 & \cellcolor{gray!30}\color{gray}0.30 \\
\midrule
\color{gray}\textbf{\textbf{Doe-1}}$^\dagger$ & \color{gray}Camera$^*$ & \color{gray}QA & \color{gray}0.37 & \color{gray}0.67 & \color{gray}1.07 & \cellcolor{gray!30}\color{gray} 0.70& \color{gray} 0.02& \color{gray} 0.14& \color{gray} 0.47& \cellcolor{gray!30}\color{gray}  0.21 \\
\bottomrule
\end{tabular}%
\label{tab:sota-plan}
\vspace{-5mm}
\end{table*}

\subsection{Implementation Details}
We use the pre-trained Lumina-mGPT 7B~\cite{liu2024lumina} to initialize our model and fine-tune the model for 5 epochs on the BDD100k~\cite{bdd100k} dataset to improve the conditioned image generation capability for driving scene images. We take as input images with a resolution of $672\times 384$ and only adopt the images from the camera. The resolution of the action tokenizer is set to 0.02m, where the scaling factor is set to 10000 for displacements and 1000 for yaw angles.

For training, we use the AdamW~\cite{loshchilov2017fixing} optimizer with a cosine learning rate scheduler. The initial learning rate is set to $1\times 10^{-5}$ with a weight decay of 0.1. To emphasize the accuracy of action prediction, we increase the loss weight for action tokens in the input sequence by a factor of 5. Z-loss is applied with a weight of $1\times 10^{-5}$ to stabilize the training. We train our model for 16 epochs on 8 A100 GPUs with a total batch size of 24.

\subsection{Close-Loop Autonomous Driving}
Figure~\ref{fig:supp_teaser} shows the visualizations of \textbf{Doe-1} for closed-loop autonomous driving. 
We model perception, planning, and prediction as the transitions of observation$\rightarrow$description, description$\rightarrow$action, and action$\rightarrow$observation, respectively.
We observe that the proposed method can correctly generate scene descriptions, answer questions about the scene, plan ego trajectory, and predict future observations correctly with one trained model without finetuning.

\subsection{Visual Question-Answering}
We use visual question-answering to evaluate the perception ability of our \textbf{Doe-1}.
We compare our method with OmniDrive~\cite{omnidrive} with 3D Q-Former (OmniDrive-3D), 2D Q-Former (OmniDrive-2D), and dense BEV (OmniDrive-BEV) on the OmniDrive-nuScenes~\cite{omnidrive} benchmark, as shown in Table~\ref{vqa}.
We use bold numbers to denote the best results.
Note that OmniDrive uses surrounding cameras as inputs, while our \textbf{Doe-1} only uses the front camera.
Still, we see that our model achieves competitive results on both visual caption and counterfactual reasoning tasks.

\textbf{Visualizations.}
We provide a qualitative analysis of the visual question-answering results in Figure~\ref{fig:vis_vqa}. 
We see that our \textbf{Doe-1} correctly describes the scene and answers the questions about the input image.

\subsection{Action-Conditioned Video Generation}
We evaluate the prediction ability of \textbf{Doe-1} on the action-conditioned video generation, where we adopt accurate actions (displacements in the BEV space) as the condition.
We compare our model with existing real-world world models in Table~\ref{gen}.
We see that \textbf{Doe-1} achieves comparable performance with Drive-WM~\cite{wang2023driving} and GenAD~\cite{yang2024generalized}, yet underperforms Vista~\cite{vista}.
Still, our model is the first to use the autoregressive architecture instead of diffusion, facilitating the joint perception and planning of \textbf{Doe-1}.

\textbf{Visualizations.}
Figure~\ref{fig:vis_gen} shows the generated sequences of images given an image and trajectory as conditions.
We see that \textbf{Doe-1} generates high-quality images following the prompt actions.
They show consistency in the 3D structure and demonstrate the ability of \textbf{Doe-1} to understand the evolutions of the 3D world.

\begin{table}[t]
\setlength{\tabcolsep}{0.01\linewidth}
\centering
\caption{\textbf{Ablation study on planning strategies.} 
``Description" means no other prompt between observations and actions. ``Mask" means the plan of preceding frames will not be masked during training and evaluation.
}
\vspace{-3.5mm}
\begin{tabular}{l|cccc|cccc}
\toprule
\multirow{2}{*}{Ablation} &
\multicolumn{4}{c|}{L2 (m) $\downarrow$} & 
\multicolumn{4}{c}{Collision Rate (\%) $\downarrow$} \\
 & 1s & 2s & 3s & \cellcolor{gray!30}Avg. & 1s & 2s & 3s & \cellcolor{gray!30}Avg.  \\
 \midrule
Description & 0.70 & 1.68 & 2.99 & \cellcolor{gray!30}1.79 & 0.07 & 0.44 & 2.69 & \cellcolor{gray!30}1.07  \\
Mask & 0.95 & 2.94 & 4.59 & \cellcolor{gray!30} 2.83 & 1.11 & 2.82 & 4.23 & \cellcolor{gray!30}2.72  \\
Full Model & \textbf{0.50} & \textbf{1.18} & \textbf{2.11} & \cellcolor{gray!30}\textbf{1.26} & \textbf{0.04} & \textbf{0.37} & \textbf{1.19} & \cellcolor{gray!30}\textbf{0.53}  \\

\bottomrule
\end{tabular}%
\label{tab:ablation}
\vspace{-6.2mm}
\end{table}

\subsection{End-to-End Motion Planning}

We evaluate the action planning performance of our \textbf{Doe-1} following existing end-to-end autonomous driving methods~\cite{genad,uniad,vad}, as shown in Table~\ref{tab:sota-plan}.
Additionally, we also compute the average performance across all previous frames for each time step at the bottom of the table following VAD~\cite{vad}.
Though our \textbf{Doe-1} does not achieve the best results, it demonstrates competitive performance with existing methods using only question-answering pairs as the auxiliary supervision.
Note that using more supervision signals generally results in better performance, with the cost of expensive annotations.
Also, our model only takes the front camera as input, while the other vision-based methods use surrounding cameras.
Still, our model plans the future trajectory with a satisfactory collision rate.
In particular, \textbf{Doe-1} delivers a small collision rate within 1 second, which is the most important factor in the practical close-loop scenario.

\subsection{Analysis}

\textbf{Effect of Different Planning Strategies.}
\textbf{Doe-1} leverages the perceived descriptions before generating the current action and masks the previous frames of the generated actions to avoid accumulation of error. 
Table~\ref{tab:ablation} demonstrates the effectiveness of our design, which shows that the planning performance is influenced by the constraints of the textual modality. 
The mask mechanism also effectively prevents significant error accumulation.

\section{Conclusion}
In this paper, we have presented a large driving world model (\textbf{Doe-1}) for closed-loop autonomous driving.
While existing end-to-end autonomous driving methods demonstrate strong planning performance, they are still open-loop and suffer from information loss with hand-crafted scene representations. 
We address this with a next-token prediction formulation and model perception, prediction, and planning with the transitions between multi-modal tokens.
We have conducted extensive experiments on the widely used nuScenes dataset and demonstrated the effectiveness of \textbf{Doe-1} on visual question-answering, action-conditioned video generation, and end-to-end motion planning.

\textbf{Limitations.} \textbf{Doe-1} only takes the front-view images as inputs due to the inefficiency of using multi-view inputs.
However, surround-view information is critical for safe autonomous driving and is an interesting future direction.

\end{document}